\documentclass[10pt,twocolumn]{article}

\usepackage{amsmath,amssymb,amsthm}
\usepackage{balance}
\usepackage{booktabs}
\usepackage{graphicx}
\usepackage{microtype}
\usepackage{multirow}
\usepackage{url}
\usepackage{xcolor}
\usepackage[hidelinks]{hyperref}

\newtheorem{proposition}{Proposition}
\newtheorem{theorem}{Theorem}
\newtheorem{definition}{Definition}
\newcommand{\scout}{\textsf{SCOUT}}
\newcommand{\auc}{\operatorname{AUC}}
\newcommand{\riauc}{\operatorname{RI\text{-}AUC}}
\newcommand{\ind}{\mathbf{1}}
\newcommand{\E}{\mathbb E}
\newcommand{\Pp}{\mathbb P}

\title{When Identical Rows Disagree:\\
From Benchmark Identifiability to Replication-Robust Anomaly Detection}
\author{Jie Deng}
\date{}

\begin{document}
\maketitle

\begin{abstract}
A released table is often treated as an i.i.d. sample, although its repeated rows may
encode business frequency, repeated entities, joins, resampling, or extraction errors.
We show that this ambiguity is one hidden measurement layer with three consequences:
feature-identical rows impose an attained evaluation ceiling, row-weighted AUROC is
sensitive to replication, and row-trained detectors learn a multiplicity-size-biased
law. An exact-row audit of all 690 OddBench datasets finds train--test overlap in 355,
feature-identical label conflict in 147, and a test anomaly identical to a training
normal in 137. Switching from row to support weighting changes AUROC by at least 0.05
on 50--61 datasets across four classical detector geometries.

We turn the same observational quotient into \scout{} (Support--Count Orthogonalized
Unsupervised Testing). \scout{} fits a one-class model once per distinct support pattern
and, only when exposure is meaningful, a separate conditional count model. Factorwise
split-conformal calibration yields marginal false-positive-rate control, while the
support channel is exactly invariant to arbitrary positive row replication. On 686
OddBench datasets and five seeds, support-only \scout{} is non-inferior to row-wise
Isolation Forest in raw AUROC (difference $-0.0026$, 95\% interval
$[-0.0057,0.0004]$) and improves replication-invariant AUROC by 0.0049. Across 100 real
datasets, 691,385 external normal-support dataset--seed evaluations track nominal false-positive levels,
and four backbones remain exactly unchanged under controlled replication. Semi-synthetic
count interventions on real covariates further expose the boundary: conditional modeling adds only
0.003--0.006 AUROC under weak rate heterogeneity but 0.057--0.083 under strong
heterogeneity. The result is a closed diagnostic-to-detector framework that states
when multiplicity is signal, nuisance, or fundamentally uninterpretable.
\end{abstract}


\section{Introduction}
\label{sec:introduction}

Suppose the same released feature vector occurs 500 times. It may describe 500
independent legitimate transactions, one transaction replicated by a join, a replay
attack, or 500 observations whose entity and time identifiers were removed. These
interpretations imply different anomaly decisions, yet a static table exposes only the
feature pattern and its multiplicity. Ordinary density-, neighbor-, and isolation-based
detectors silently let multiplicity reshape feature geometry. Global deduplication
makes the opposite silent decision by discarding all frequency evidence.

The problem begins before model fitting. If two feature-identical test rows carry
different labels, every deterministic detector using the released features must tie
them. If one such row is replicated, ordinary AUROC gives that unresolved class more
pairwise weight. Thus a benchmark can simultaneously withhold information needed to
separate its labels and amplify that ambiguity through row frequency. A detector paper
that changes only its encoder or prediction head cannot recover missing provenance,
entity, time, or exposure information.

We develop a single framework around the \emph{observational quotient}: exact equality
partitions a released table into distinct support patterns and their multiplicities.
The quotient first acts as a certificate. It yields an exact AUROC decomposition, an
attained finite-sample ceiling for every deterministic feature-measurable score, and a
replication-invariant sensitivity metric. It then acts as a model interface. The
expanded row law is a multiplicity-size-biased version of the support law, motivating a
factorized detector that keeps geometry and conditional count separate. The primary
intervention is therefore a change in statistical unit---from expanded rows to observed
support plus an explicit count measurement---rather than another backbone.

The audit makes this more than a corner case. Among all 690 public OddBench datasets,
355 official splits share exact rows across train and test; 147 contain identical
features with conflicting labels; and, in 137, at least one test anomaly is identical to
a released training normal. Across Isolation Forest, ECOD, HBOS, and COPOD, changing
only the evaluation unit from rows to observed support changes AUROC by at least 0.05 on
50--61 datasets and changes the top-ranked model on 30. We do not infer that every
duplicate is erroneous. The finding is narrower and more actionable: multiplicity
semantics vary materially, so neither training nor evaluation should leave them
implicit.

We introduce \scout{}, a backbone-agnostic one-class detector. One representative per
normal equivalence class enters a structural model; a separate XGBoost regressor models
exposure-normalized log count conditional on that representative. Held-out
support-class calibration produces a structural p-value $p_S$ and, when exposure is
comparable, a frequency p-value $p_M$. Support-only scoring is the safe default.
Bonferroni combination is enabled only under explicit exposure semantics. This
architecture does not claim to tell fraud from popularity or pipeline failure: we prove
that business and artifact intensities are not separately identifiable when only their
product is observed.

\begin{figure*}[t]
\centering
\includegraphics[width=0.98\textwidth]{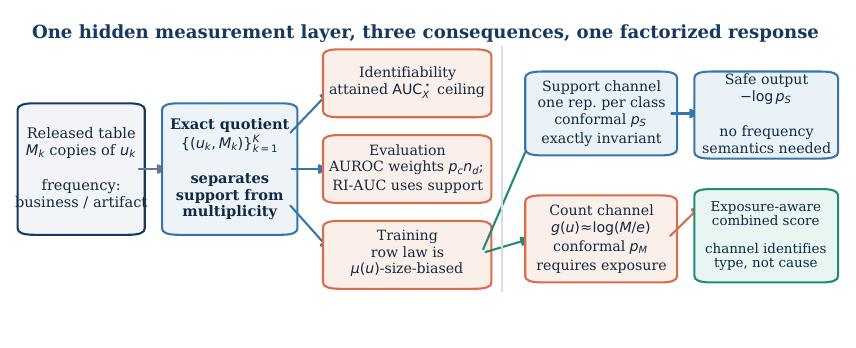}
\caption{The fused thesis. A released row table is an expanded observation of support
patterns $u_k$ and multiplicities $M_k$. The quotient diagnoses information and metric
limits, explains training size bias, and supplies the interface for \scout{}. The
support output needs no frequency semantics; the combined output requires comparable
exposure and attributes anomaly type, not its unobserved cause.}
\label{fig:overview}
\end{figure*}

Figure~\ref{fig:overview} shows the unified argument. The contributions are:
\begin{itemize}
    \item \textbf{Certificate.} We derive an exact equivalence-class decomposition of
    AUROC, a sharp attained feature-identifiability ceiling, and the canonical
    replication-invariant member of a natural separable pair-weight family.
    \item \textbf{Measurement model.} We prove that row sampling is multiplicity-size-
    biased and that business versus artifact frequency is non-identifiable without
    auxiliary information. These results connect benchmark disagreement to detector
    training and delimit every frequency claim.
    \item \textbf{Detector.} We propose \scout{}, whose support scores are exactly
    invariant to arbitrary positive replication and whose factorwise conformal tests
    have finite-sample marginal false-positive control under stated exchangeability.
    \item \textbf{Evidence and boundary.} We combine a complete 690-dataset audit,
    four-detector evaluation sensitivity, 3,430 matched full-benchmark fits, mechanism
    simulations, external real-support calibration, four-backbone interventions, and
    real-covariate semi-synthetic count tests. Negative results and failure conditions
    are reported rather than hidden.
\end{itemize}

\section{Related Work and Novelty Boundary}
\label{sec:related}

\paragraph{Tabular anomaly detection and benchmarks.}
ADBench compares 30 methods on 57 datasets~\cite{han2022adbench}; an independent JMLR
study evaluates 33 algorithms on 52 real datasets~\cite{bouman2024howmany}; and
MacrOData expands evaluation to 2,446 curated datasets~\cite{ding2026macrodata}.
Recent methods learn random deep isolation representations~\cite{xu2023dif}, masked
feature dependencies~\cite{yin2024mcm}, cross-sample interactions
~\cite{thimonier2024npt}, decomposed embeddings~\cite{ye2025drl}, or stabilized
densities~\cite{rozner2024vsde}. These improve representations or scoring geometries;
they do not specify whether repeated rows are independent observational units or an
expansion artifact.

\paragraph{Data quality and evaluation.}
TabReD documents split and protocol pitfalls in tabular deep learning
~\cite{rubachev2025tabred}. SelfClean uses intrinsic representations to rank duplicates,
label errors, and off-topic samples~\cite{groger2024selfclean}. Our scope is different:
we neither auto-delete duplicates nor assume they are errors. We derive what identical
released features make unidentifiable, how multiplicity weights a rank metric, and how
to construct a detector invariant to declared nuisance replication. Work on AUROC under
imbalance clarifies metric behavior under changing class prevalence
~\cite{mcdermott2024auroc}; our transformation changes weights \emph{within} each label
through equivalence-class replication while keeping feature support fixed.

\paragraph{Robustness and conformal novelty detection.}
Invariant anomaly detection regularizes across environments to resist distribution
shift~\cite{carvalho2023invariant}. Our nuisance is an algebraic transformation within
exact-feature classes, permitting exact rather than distributional invariance. Split
conformal inference converts any held-out nonconformity score into a marginally valid
test~\cite{vovk2005algorithmic}; conformal e-values extend novelty detection to FDR
control~\cite{bashari2023derandomized}. Recent robust conformal outlier detection studies
contaminated reference sets and active cleaning~\cite{bashari2025robust}. \scout{} does
not claim conformal ranks, Bonferroni correction, or contamination robustness as new;
its contribution is the support--count sampling unit and replication orthogonality.

\paragraph{Semantic and frequency anomalies.}
CADES combines complementary conformal scores for continuous-time event sequences
~\cite{zhang2025cades}. EventADL separates semantic and frequency rules for cloud
events~\cite{pham2026eventadl}. These methods assume event time or system structure. Our
setting is a static released table in which that context may be absent. The quotient
therefore exposes an impossibility boundary and provides a support-only fallback rather
than inventing event semantics.

\paragraph{Foundation and graph features.}
AnoLLM serializes mixed-type rows~\cite{tsai2025anollm}; FoMo-0D and OUTFORMER learn
zero-shot outlier inference from synthetic pretraining
~\cite{shen2025fomo,ding2026outformer}; and TFM4GAD adds graph-derived features before
tabular foundation inference~\cite{liu2026tfm4gad}. A pretrained or graph encoder is
valuable when text, entities, edges, or time add information. It can also replace the
structural backbone in \scout{}. But any deterministic encoder of the same released row
preserves observational equivalence and cannot identify why that row was replicated.
This is why a pretrained-feature-plus-XGBoost pipeline was not retained as the central
contribution.

\section{The Observational-Equivalence Certificate}
\label{sec:certificate}

Let the labeled evaluation sample be
$D=\{(x_i,y_i)\}_{i=1}^n$, with $y_i\in\{0,1\}$. Define $i\sim j$ iff
$x_i=x_j$ in the released representation, and let $\mathcal C$ be the resulting classes.
For $c\in\mathcal C$, let $p_c$ and $n_c$ be its positive and negative row counts,
$P=\sum_c p_c$, $N=\sum_c n_c$, and $s_c=f(x_c)$ the score of any deterministic
feature-measurable detector, conditioning on any fitted model and random seed. Write
$L(a,b)=\ind[a>b]+\tfrac12\ind[a=b]$.

\subsection{Exact decomposition and an attained ceiling}

\begin{proposition}[Equivalence-class AUROC]
For every deterministic feature-measurable score,
\begin{equation}
\auc(f;D)=\frac{1}{PN}\sum_{c,d\in\mathcal C}p_c n_d L(s_c,s_d).
\label{eq:aucdecomp}
\end{equation}
The within-class contribution $(2PN)^{-1}\sum_c p_cn_c$ is fixed.
\end{proposition}

Equation~\eqref{eq:aucdecomp} isolates two distinct facts. Conflicting labels in one
class must tie, creating an information limit. Multiplicity also changes cross-class
weights $p_cn_d$, so duplicating already observed row--label pairs can change AUROC
without changing either the detector scores or feature support.

Define the empirical anomaly fraction $q_c=p_c/(p_c+n_c)$.

\begin{proposition}[Attained feature-identifiability ceiling]
Ranking classes by nonincreasing $q_c$ maximizes empirical AUROC over every
deterministic score measurable from the released features. Hence
\begin{equation}
\auc_X^\star(D)=\auc\bigl(y_i,q_{[i]}\bigr)
\label{eq:ceiling}
\end{equation}
is a sharp, attained finite-sample ceiling, where $[i]$ is row $i$'s class.
\end{proposition}

The adjacent-swap proof is in Appendix~\ref{app:proofs}. The ceiling uses test labels
only as a post hoc certificate; it is neither an unsupervised detector nor a quantity
for model selection, and it makes no population-generalization claim.

\subsection{A replication-invariant sensitivity estimand}

Let $\widetilde p_c=\ind[p_c>0]$, $\widetilde n_c=\ind[n_c>0]$,
$\widetilde P=\sum_c\widetilde p_c$, and
$\widetilde N=\sum_c\widetilde n_c$.

\begin{definition}[Replication-invariant AUROC]
\begin{equation}
\riauc(f;D)=
\frac{1}{\widetilde P\widetilde N}
\sum_{c,d}\widetilde p_c\widetilde n_dL(s_c,s_d).
\label{eq:riauc}
\end{equation}
Operationally, evaluation occurs once per observed (class, label) pair. A mixed-label
class contributes one tied positive and one tied negative support point.
\end{definition}

\begin{proposition}[Invariance and canonical separable weights]
RI-AUC is unchanged when any already observed row--label pair is replicated an
arbitrary positive number of times. Moreover, within normalized pairwise rank metrics
whose nonnegative class weights separate as $w^+(p_c)w^-(n_d)$, invariance to arbitrary
positive replication for every dataset and score forces $w^+$ and $w^-$ to be constant
on the positive integers. With label-wise normalization this is Eq.~\eqref{eq:riauc}.
\end{proposition}

RI-AUC is therefore not proposed as a universally superior replacement. Ordinary
AUROC answers the row-frequency-weighted question and is appropriate when each row is
the intended unit. RI-AUC answers a support-weighted sensitivity question when
multiplicity may be nuisance. Their signed difference
$\Delta_{\rm rep}=\riauc-\auc$ measures multiplicity leverage, not automatically bias.

\section{Multiplicity as a Hidden Measurement Layer}
\label{sec:measurement}

For one-class training, let $U=\{u_k\}_{k=1}^K$ be the distinct normal support and
$M_k=\sum_i\ind[X_i=u_k]$ its multiplicity in a stated exposure window. At population
level, suppose observational classes $(U_k,M_k)$ are i.i.d., with $U\sim Q$ and
$\mu(u)=\E[M\mid U=u]$. Define $P_{\rm row}$ as the size-biased law obtained by
selecting an expanded row from this population (equivalently, the large-batch limit of
uniform row selection).

\begin{proposition}[Row-law size bias]
If $0<\E_Q\mu(U)<\infty$, the law $P_{\rm row}$ of a uniformly selected expanded row
satisfies
\begin{equation}
\frac{dP_{\rm row}}{dQ}(u)=\frac{\mu(u)}{\E_Q[\mu(U)]}.
\label{eq:sizebias}
\end{equation}
\end{proposition}

Thus an ordinary row-density model does not estimate support geometry $Q$ alone. It
estimates a mixture of support prevalence and conditional count. The same multiplicity
that weights evaluation in Eq.~\eqref{eq:aucdecomp} weights representation learning and
model fitting in Eq.~\eqref{eq:sizebias}.

Suppose exposure $e>0$ is known and mean multiplicity factors as
\begin{equation}
\mu(u,e)=e\,b(u)\,a(u),
\label{eq:factor}
\end{equation}
where $b$ represents business/event intensity and $a$ an artifact or replication
mechanism.

\begin{proposition}[Frequency-cause non-identifiability]
From observations $(U,M,e)$ alone, $b$ and $a$ in Eq.~\eqref{eq:factor} are not
separately identifiable without restrictions or auxiliary metadata.
\end{proposition}

For any positive function $h$, replacing $b$ by $bh$ and $a$ by $a/h$ preserves every
observable mean. Repeated windows, provenance, entity identifiers, or interventions are
needed to assign cause. This impossibility result dictates the method interface:
separate support from count, expose both evidence channels, and disable causal language.

\section{SCOUT: From Certificate to Detector}
\label{sec:method}

\scout{} reuses the quotient rather than treating the audit as a preprocessing report.
Normal support classes are deterministically split into proper-training indices
$I_{\rm fit}$ and calibration indices $I_{\rm cal}$ after quotienting, so replication
cannot alter the partition.

\subsection{Orthogonal support and count channels}

The \textbf{support channel} fits a scaler and any one-class backbone $A_S$ on
$\{u_k:k\in I_{\rm fit}\}$ with one unit of weight per class. Larger
$r_S(u)=A_S(u)$ is more anomalous. The calibration multiset is
$R_S=\{r_S(u_k):k\in I_{\rm cal}\}$.

When comparable exposure is available, the \textbf{count channel} fits
$g_\theta(u)$ to
\begin{equation}
z_k=\log(M_k/e),\qquad k\in I_{\rm fit}.
\end{equation}
For two-sided frequency anomalies,
\begin{equation}
r_M(u,M,e)=|\log(M/e)-g_\theta(u)|.
\label{eq:countscore}
\end{equation}
One-sided high or low residuals are used only when the application fixes a direction.
Conditioning matters when expected rate depends on covariates; Section~\ref{sec:results}
also reports the boundary where that dependence is too weak to justify the extra model.

For channel $j\in\{S,M\}$ and test nonconformity $r_j^\star$, the conservative
split-conformal p-value is
\begin{equation}
p_j=\frac{1+\sum_{k\in I_{\rm cal}}\ind[r_{j,k}\geq r_j^\star]}
{1+|I_{\rm cal}|}.
\label{eq:pvalue}
\end{equation}
The safe default score is $-\log p_S$. With scientifically comparable exposure, we
report both channels and optionally combine them as
\begin{equation}
p_{\rm comb}=\min\{1,2\min(p_S,p_M)\},\qquad
s_{\rm comb}=-\log p_{\rm comb}.
\label{eq:combine}
\end{equation}
The smaller factor p-value attributes the statistical type of surprise, not its cause.

\begin{theorem}[Exact replication orthogonality]
Fix the seed and deterministic representative order. Replicating any training or
scoring row a positive integer number of times leaves every support nonconformity,
support p-value, and support anomaly score exactly unchanged.
\label{thm:invariance}
\end{theorem}

\begin{theorem}[Factorwise and combined FPR control]
Condition on the proper-training data. If a null test support class and support
calibration classes are exchangeable, then
$\Pp(p_S\leq\alpha)\leq\alpha$. If exposure-normalized count residuals are also
exchangeable, the same holds for $p_M$ and
\begin{equation}
\Pp(p_{\rm comb}\leq\alpha)\leq\alpha
\end{equation}
without assuming channel independence.
\label{thm:fpr}
\end{theorem}

The proofs are elementary but operationally important: Theorem~\ref{thm:invariance}
is algebraic, whereas Theorem~\ref{thm:fpr} depends on the sampling unit and
exchangeability. It is marginal class-level FPR control, not row-wise batch FDR or a
guarantee under temporal drift.

\paragraph{Implementation.}
We reserve 20\% of support classes for calibration, with at least 20 fit and 20
calibration classes. Primary experiments use Isolation Forest~\cite{liu2008iforest} with 200 trees and at
most 256 support samples per tree. The conditional count model is XGBoost with 160
depth-3 trees, learning rate 0.05, and row/column subsampling 0.8; constant training
counts invoke a constant regressor. XGBoost is a flexible working regression model, not
a claimed count likelihood; Poisson-deviance, negative-binomial, or application-specific
rate models can replace it without changing the factorization. The API returns support-only output unless frequency
is explicitly enabled with exposure. Storage is $O(Kd)$ plus the backbone cost, where
$K\leq n$.

\section{Study Design}
\label{sec:protocol}

We test five linked questions: (RQ1) Is the hidden measurement layer material in a
complete public benchmark? (RQ2) Do the two channels recover their intended mechanisms?
(RQ3) Do conformal support tests calibrate beyond Gaussian simulation? (RQ4) Does
replication robustness preserve real-data utility? (RQ5) Is invariance causal under a
replication intervention and independent of the structural backbone?

\subsection{Complete audit and evaluation sensitivity}

OddBench provides 690 one-class datasets with released normal training and labeled test
splits~\cite{ding2026macrodata}. We audit all released arrays using bitwise exact numeric
rows after canonicalizing signed zero; no rounding, distance threshold, or learned
representation defines equivalence. Diagnostics include exact train--test overlap,
within-test label conflict, test anomalies equal to training normals, the ceiling in
Eq.~\eqref{eq:ceiling}. A separate appendix output reports held-out single-feature AUROC
as a marginal-easiness warning; it is not part of the multiplicity thesis and is not
called leakage because feature availability at the original decision time is unknown.

Isolation Forest, ECOD, HBOS, and COPOD are trained on official normal rows using a
shared implementation~\cite{zhao2019pyod}. We compare ordinary AUROC with RI-AUC for
every detector--dataset pair. Dataset-level bootstrap intervals use 50,000 resamples;
paired Wilcoxon tests are Holm-adjusted across four detectors. We also compare detector
rankings under both metrics.

\subsection{Mechanism identification and frequency boundary}

Controlled normal supports are Gaussian, with nonlinear conditional counts
$1+\operatorname{Poisson}(\exp f(u))$. Support anomalies shift geometry; frequency
anomalies retain normal features but receive an abnormal conditional count boost;
``both'' combines mechanisms; and ``mixed'' evaluates their union. Dimensions are
4, 8, and 16 with seeds 11, 29, 47, 83, and 101, giving 15 settings. Baselines are
row-wise Isolation Forest, deduplicated Isolation Forest, count-augmented Isolation
Forest, and unconditional frequency. Labels select only the reported mechanism-specific
ablation, never a fitted model or hyperparameter.

A revision-stage stress test, fixed before its outcomes were inspected, raises the
dimension to 50 and 100 with the same five seeds. It evaluates the original conditional
log-rate function and a stronger version obtained by multiplying its log-rate signal by
1.5. This extension is reported separately rather than retroactively reweighting the
primary 15 settings.

To stress the count model on non-Gaussian geometry, we select the same 100 OddBench
datasets used for the replication intervention, selected by training-duplicate-rate
strata without test-label performance, and extract genuine normal support vectors,
and split support 70/30. Counts and burst labels are deliberately semi-synthetic because
OddBench has no window exposure. A fixed nonlinear projection produces either weak
($0.35$) or strong ($1.0$) covariate rate heterogeneity. Training and test exposure are
1 and 2; 10\% of test support receives a rate boost of 1.5, 2, or 4. Each of the six
\scout{}-frequency versus unconditional-frequency comparisons uses datasets as paired
units, a one-sided Wilcoxon test, Holm correction, and 50,000 bootstrap resamples.

\subsection{Calibration, utility, and intervention}

Synthetic validity uses 15 fitted models and 12 independent null batches of 400 support
classes per fit, for 72,000 null tests per channel at
$\alpha\in\{0.01,0.05,0.10\}$. External support calibration uses 100 real datasets,
five seeds, an internal proper/calibration split, and a disjoint held-out normal-support
set. It yields 500 dataset--seed runs and 691,385 dataset--seed support evaluations at each alpha. The micro
proportion is descriptive under within-dataset dependence; we also bootstrap the 500
run-level rates.

Four OddBench datasets have fewer than 40 distinct normal support points, leaving 686
for full utility evaluation. \scout{} and row-wise Isolation Forest use five matched
seeds. We prespecify a $-0.01$ non-inferiority margin for mean raw AUROC and report both
raw and RI-AUC. The dataset is the inferential unit.

For the intervention, 100 datasets are stratified across training duplicate rate. We
select 5\% of normal support classes and append replicas totaling 50\% of the original
training-row count, then independently refit models. We record pointwise score change,
score-rank Spearman correlation, and AUROC change. The support construction is repeated
with Isolation Forest, ECOD, HBOS, and COPOD.

\section{Results}
\label{sec:results}

\subsection{RQ1: disagreement is common and changes conclusions}

Table~\ref{tab:audit} reports the complete audit. Exact train--test overlap occurs in
51.4\% of datasets, and 21.3\% contain a feature-identical label conflict somewhere in
the released evaluation data. Seventeen datasets have an attained feature ceiling below
0.95 and seven below 0.90. These counts are not error labels: an intentional repeated
entity can be scientifically valid. They certify that the released row alone is
insufficient for some labels and that source semantics must determine the unit.

\begin{table}[t]
\caption{Exact-row audit of all 690 OddBench datasets.}
\label{tab:audit}
\centering
\footnotesize
\begin{tabular}{lrr}
\toprule
Diagnostic & Count & Percent \\
\midrule
Train--test exact overlap & 355 & 51.4 \\
Any identical-feature label conflict & 147 & 21.3 \\
Test-internal label conflict & 140 & 20.3 \\
Test anomaly equals training normal & 137 & 19.9 \\
Ceiling $\auc_X^\star<.99/.95/.90$ & 29/17/7 & 4.2/2.5/1.0 \\
\bottomrule
\end{tabular}
\end{table}

The evaluation effect is detector-general (Table~\ref{tab:metric}). Mean RI-AUC is
lower because collision-heavy classes lose row leverage, but the signed average does
not describe individual instability: 50--61 datasets per detector have
$|\Delta_{\rm rep}|\geq0.05$. Across 2,760 detector--dataset cells, raw and RI rankings
have Kendall $\tau=0.929$; nevertheless, 64 datasets have at least one rank change and
30 change their top detector. For Isolation Forest, test-anomaly/training-normal
collision rate correlates with published AUROC at $\rho=-0.221$
($p=4.76\times10^{-9}$). This is association, not a causal leakage estimate.

\begin{table}[t]
\caption{Evaluation sensitivity. $\Delta$ is RI-AUC minus raw AUROC; CIs bootstrap
datasets. All Holm-adjusted $p\leq3.22\times10^{-4}$.}
\label{tab:metric}
\centering
\scriptsize
\begin{tabular}{@{}lccc@{}}
\toprule
Detector & Raw/RI mean & $\Delta$ (95\% CI) & $|\Delta|\geq.05$ \\
\midrule
IForest & .7119/.7007 & $-.0112\;[-.0153,-.0076]$ & 61 \\
ECOD    & .6650/.6595 & $-.0054\;[-.0090,-.0019]$ & 60 \\
HBOS    & .6732/.6677 & $-.0055\;[-.0082,-.0029]$ & 50 \\
COPOD   & .6539/.6470 & $-.0070\;[-.0112,-.0026]$ & 60 \\
\bottomrule
\end{tabular}
\end{table}

\begin{figure*}[t]
\centering
\includegraphics[width=0.95\textwidth]{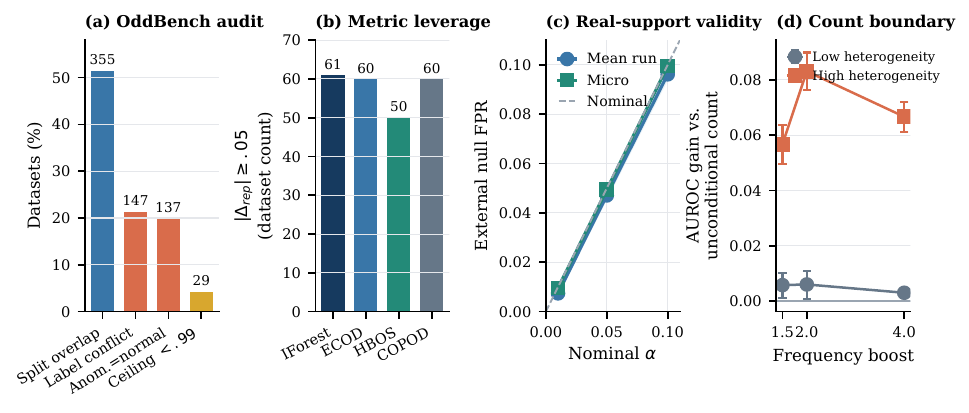}
\caption{Evidence for materiality, validity, and the frequency boundary. (a) Complete
benchmark prevalence. (b) Datasets with at least 0.05 absolute row-versus-support AUROC
change. (c) External held-out normal-support FPR: micro proportions and mean
dataset--seed rates, with run-level bootstrap bands. (d) Conditional-frequency AUROC
gain over an unconditional count score on real support geometry; bars are paired
dataset bootstrap intervals.}
\label{fig:fused_evidence}
\end{figure*}

\subsection{RQ2: factorization recovers mechanisms, not every advantage}

Table~\ref{tab:synthetic} separates four targets. The support channel matches
deduplicated Isolation Forest on support anomalies (.9854 versus .9852). The conditional
frequency channel reaches .9946 on frequency anomalies, versus .9524 for unconditional
frequency and .7601 for count-augmented Isolation Forest. On mixed anomalies, the
combined score reaches .9847 versus .9133 for count-augmented Isolation Forest; the
paired difference is .0714 with interval $[.0599,.0828]$, wins in all 15 settings, and
has $p=6.10\times10^{-5}$.

\begin{table*}[t]
\caption{Mean AUROC across five seeds and three dimensions. SCOUT-S, -M, and -C are
support, frequency, and combined channels. Bold marks the prespecified channel for the
target, not label-based model selection.}
\label{tab:synthetic}
\centering
\small
\begin{tabular}{lrrrrrrr}
\toprule
Target & Row IF & Dedup IF & Count-aug. IF & Uncond. freq. & SCOUT-S & SCOUT-M & SCOUT-C \\
\midrule
Support   & .9859 & .9852 & .9867 & .5628 & \textbf{.9854} & .5351 & .9707 \\
Frequency & .5086 & .5009 & .7601 & .9524 & .5012 & \textbf{.9946} & .9925 \\
Both      & .9839 & .9832 & .9931 & .9481 & .9833 & .9850 & \textbf{.9910} \\
Mixed     & .8261 & .8231 & .9133 & .8211 & .8233 & .8382 & \textbf{.9847} \\
\bottomrule
\end{tabular}
\end{table*}

The combination is .0146 below support-only on pure support anomalies. This multiple-
testing cost falsifies a policy of always combining channels and supports the
support-only default.

\begin{table*}[t]
\caption{Revision-stage dimensional stress test. Entries are mean paired AUROC
differences with 95\% bootstrap intervals over seed--dimension settings. The first two
columns favor the prespecified SCOUT channel; the last exposes the cost of reserving
20\% of support classes for calibration.}
\label{tab:highdim}
\centering
\small
\begin{tabular}{lrrrr}
\toprule
Regime & Settings & Mixed: C $-$ count-aug. & Frequency: M $-$ uncond. & Support: S $-$ dedup \\
\midrule
$d=4,8,16$, rate scale 1.0 & 15 & $.0714\;[.0599,.0828]$ & $.0422\;[.0365,.0476]$ & $.0002\;[-.0021,.0038]$ \\
$d=50,100$, rate scale 1.0 & 10 & $.1532\;[.1367,.1713]$ & $.0415\;[.0371,.0462]$ & $-.0217\;[-.0351,-.0094]$ \\
$d=50,100$, rate scale 1.5 & 10 & $.1700\;[.1524,.1901]$ & $.0916\;[.0856,.0981]$ & $-.0246\;[-.0460,-.0083]$ \\
\bottomrule
\end{tabular}
\end{table*}

Table~\ref{tab:highdim} shows that the frequency factorization does not rely on a
low-dimensional feature space: both high-dimensional regimes win in all ten settings
for the mixed and frequency contrasts. The same test also identifies a failure boundary.
On pure support anomalies, split calibration loses power relative to fitting
deduplicated Isolation Forest on every support point. The loss is consistent with the
cost of withholding calibration classes when structural estimation is difficult; we do
not claim high-dimensional support-power parity.

Table~\ref{tab:frequency} tests whether conditional count modeling survives real
covariate geometry. Under weak heterogeneity, statistically detectable gains are only
.0031--.0061 AUROC. Under strong heterogeneity they rise to .0567--.0831, with 92--99
dataset wins out of 100. All six Holm-adjusted tests reject no gain, but effect size---not
the p-value---determines whether the added frequency model is useful. In every setting,
geometry-only and support scores remain near chance because anomalies were constructed
only through count; the combined test is below the frequency channel because it spends
error budget on an uninformative support channel.

\begin{table*}[t]
\caption{Real-support/semi-synthetic frequency interventions. Gain is SCOUT-M minus
unconditional frequency; CIs bootstrap 100 paired datasets. Counts and labels are
semi-synthetic, not real fraud outcomes.}
\label{tab:frequency}
\centering
\small
\begin{tabular}{rrrrrrr}
\toprule
Rate heterogeneity & Boost & Uncond. mean & SCOUT-M mean & Gain (95\% CI) & Wins & Holm $p$ \\
\midrule
.35 & 1.5 & .6898 & .6956 & $.0059\;[.0013,.0102]$ & 70 & $3.78\times10^{-4}$ \\
.35 & 2.0 & .8178 & .8239 & $.0061\;[.0007,.0108]$ & 72 & $3.78\times10^{-4}$ \\
.35 & 4.0 & .9806 & .9837 & $.0031\;[.0011,.0049]$ & 79 & $1.06\times10^{-6}$ \\
1.00 & 1.5 & .6500 & .7067 & $.0567\;[.0497,.0636]$ & 92 & $9.48\times10^{-17}$ \\
1.00 & 2.0 & .7415 & .8246 & $.0831\;[.0762,.0899]$ & 98 & $1.24\times10^{-17}$ \\
1.00 & 4.0 & .9105 & .9772 & $.0667\;[.0613,.0721]$ & 99 & $1.24\times10^{-17}$ \\
\bottomrule
\end{tabular}
\end{table*}

\subsection{RQ3: support calibration transfers to real geometry}

In the synthetic null study, combined FPR is .0050, .0455, and .0941 at nominal
.01, .05, and .10. Support and count channels are likewise at or below nominal on
average. More importantly, Figure~\ref{fig:fused_evidence}c evaluates unseen support
drawn from real OddBench normal references. Across 691,385 external tests per alpha,
micro FPR is .00958, .04936, and .09949. Mean dataset--seed FPR is .00738, .04707, and
.09615, with run-level bootstrap intervals $[.00683,.00795]$,
$[.04512,.04903]$, and $[.09363,.09867]$. The pooled proportions are more precise but
descriptive because the same datasets contribute across seeds and support points within
one dataset are dependent. This experiment
validates only the support channel; OddBench provides no exposure with which to validate
real count calibration.

\subsection{RQ4: robustness preserves full-benchmark utility}

Table~\ref{tab:utility} summarizes 3,430 matched fits. Mean raw AUROC falls by .00258,
and the entire bootstrap interval is above the prespecified $-.01$ margin. RI-AUC
improves by .00487 with a positive interval. We make no state-of-the-art claim: the
comparison isolates the cost and benefit of changing the observational unit while
holding the backbone family fixed.

\begin{table*}[t]
\caption{Five-seed, dataset-mean comparison on 686 OddBench datasets. Differences are
SCOUT support minus row-wise Isolation Forest; CIs bootstrap datasets.}
\label{tab:utility}
\centering
\small
\begin{tabular}{lrrrrrr}
\toprule
Metric & SCOUT mean & Row IF mean & Mean difference & 95\% CI & Median difference & Paired $p$ \\
\midrule
Raw AUROC & .70885 & .71142 & -.00258 & $[-.00571,.00040]$ & -.00006 & .369 \\
RI-AUC    & .70616 & .70129 &  .00487 & $[ .00236,.00768]$ &  .00019 & .016 \\
\bottomrule
\end{tabular}
\end{table*}

Median wall-clock time per \scout{} fit is .387 seconds (95th percentile .814) versus
.333 seconds (.459) for row-wise Isolation Forest on the four shared additional seeds
and the same machine. This is descriptive rather than a hardware-independent complexity
claim.

The trade-off concentrates where the measurement layer is strongest. In the highest
training-duplication quartile (172 datasets, mean duplicate-row rate .418), raw AUROC is
.0104 lower while RI-AUC is .0198 higher. In the first three quartiles, mean raw and RI
differences lie between $-.0006$ and $+.0002$. This pattern is consistent with size
bias: quotient training matters little when $K\approx n$ and changes the target most
when a few support patterns carry high multiplicity.

\begin{figure*}[t]
\centering
\includegraphics[width=0.98\textwidth]{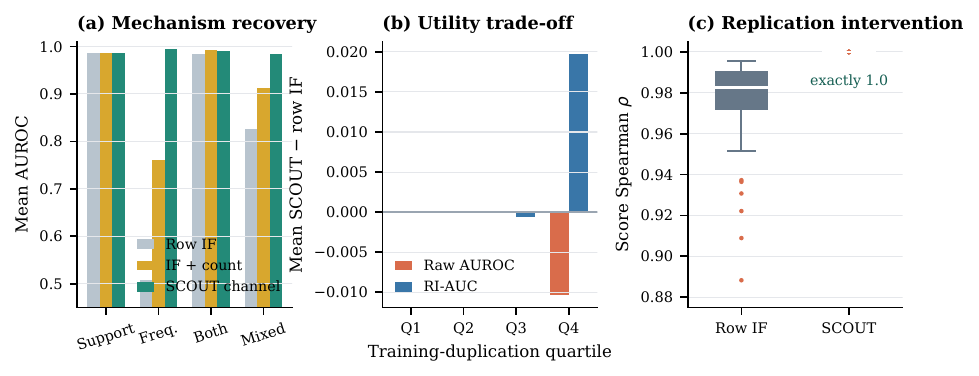}
\caption{Detector evidence. (a) Mechanism-controlled power; the SCOUT channel is
prespecified by the target. (b) Full-benchmark mean SCOUT-minus-row-IF utility by
training-duplication quartile. (c) Test-score rank stability after the controlled
training-row replication intervention.}
\label{fig:detector}
\end{figure*}

\subsection{RQ5: intervention verifies a backbone-independent guarantee}

After appending arbitrary normal replicas on 100 datasets, row-wise Isolation Forest
has median score-rank correlation .9825, minimum .8883, and six datasets below .95.
Median absolute AUROC change is .0061, its 95th percentile .0317, and maximum .0758.
Mean signed change is near zero, showing why an average-only robustness check misses
local instability.

For \scout{}, maximum pointwise support-score change is exactly zero and minimum rank
correlation is 1.0 to numerical precision for Isolation Forest, ECOD, HBOS, and COPOD.
Their mean raw/RI
AUROC pairs on the 100 datasets are .7125/.7140, .6631/.6651, .6500/.6523, and
.6433/.6448, respectively. Exact stability therefore comes from quotient construction,
not a special property of Isolation Forest; weaker backbones remain weaker.

\section{Discussion}
\label{sec:discussion}

\paragraph{What the fusion changes.}
An audit-only paper can reveal that identical rows disagree but leaves practitioners
with no detector. A detector-only paper can look like deduplication plus a count feature.
The joint theory removes both weaknesses. The same equivalence classes (i) state what
released features cannot distinguish, (ii) reveal how evaluation assigns leverage,
(iii) identify the size-biased training law, and (iv) define the nuisance-invariant
structural sample. The novelty is this closed measurement-to-method loop, not any single
off-the-shelf component.

\paragraph{When to enable frequency.}
Frequency belongs in the target when observations come from comparable windows,
exposure is measured, and abnormal rate is operationally meaningful---transactions per
active account-day, requests per service-minute, or actions per eligible user. It should
be disabled when count may result from undocumented joins, retransmission, resampling,
or benchmark construction. Even with exposure, Table~\ref{tab:frequency} shows that a
conditional model adds little when normal rate barely depends on covariates. A simple
unconditional count test may then be preferable.

\paragraph{Why neither global deduplication nor richer features resolves the problem.}
Global deduplication provides support robustness but destroys legitimate rate evidence.
Appending raw count to a detector lets count distort geometry and underperforms the
factorized channel in our mechanism experiments. A language model, tabular foundation
model, graph embedding, or Deep Isolation Forest can enrich $A_S$ when additional
context exists. If it receives only the same feature vector, however, it maps identical
rows identically and cannot infer the missing factorization in Eq.~\eqref{eq:factor}.
SCOUT is compatible with richer backbones but requires the sampling-unit decision first.

\paragraph{Limitations and external validity.}
Exact equality misses near-duplicates, quantization aliases, and repeated entities with
changing fields. Approximate equivalence needs a domain metric and can wrongly merge
distinct states, so none of our guarantees is silently extended to it. The empirical
ceiling uses evaluation labels and does not predict future generalization. RI-AUC is a
sensitivity estimand, not a universal replacement for row AUROC. Conformal validity
requires exchangeable normal support classes; temporal drift, entity dependence, or
contaminated calibration may violate it, and robust contaminated-reference methods are
complementary~\cite{bashari2025robust}. No analogous sharp average-precision ceiling is
claimed. The count channel detects unexpected conditional
rate but cannot name its cause. Its power evidence is synthetic or semi-synthetic because
the public benchmark lacks exposure metadata; a prospective exposure-labeled fraud or
event deployment remains necessary. OddBench is broad but is not a dedicated financial-
fraud benchmark. Finally, full-benchmark predictive comparison uses Isolation Forest;
the other classical backbones test causal invariance and utility on a stratified
100-dataset subset. A matched Deep Isolation Forest extension on that subset is
inconclusive rather than evidence of predictive transfer.

\paragraph{Negative directions.}
To avoid inferring strong-backbone utility from a theorem about invariance, we compared
row-wise and support-wise Deep Isolation Forest on the frozen 100-dataset subset using
its default 50-representation ensemble. Mean raw AUROC changed by $-.0036$ with interval
$[-.0166,.0059]$, so the $-.01$ non-inferiority criterion was not met; RI-AUC changed by
$-.0017$ with interval $[-.0145,.0074]$. We therefore report the result but do not
promote it as a successful backbone extension. A random-projection hypersphere detector looked strong in simulation but
reached .679 on a 20-dataset real pilot, below quotient Isolation Forest at .705. These
routes were rejected rather than promoted through selective experiments. Pretrained
feature engineering was retained as an optional backbone direction, not a solution to
an observation mechanism it cannot identify.

\section{Conclusion}

Identical rows that disagree are not merely a cleaning nuisance. They expose a hidden
measurement layer governing feature identifiability, metric weighting, and detector
training. The observational quotient makes that layer explicit. Its certificate gives a
sharp evaluation ceiling and a replication-invariant sensitivity estimand; its
population model reveals multiplicity size bias and a causal non-identifiability limit;
and its detector implementation separates stable support evidence from exposure-aware
count evidence. Complete-benchmark audits, full-scale matched fits, external
calibration, mechanism tests, and causal replication interventions support the framework
while identifying where it should not be used. The practical recommendation is simple:
declare the observational unit before optimizing the anomaly detector.

\paragraph{Reproducibility.}
The repository contains source code, fixed seeds, all successful and failed experiment
rows, unit tests, audit tables, bootstrap outputs, and figure builders. Four excluded
utility datasets---BudgetGoal, GameItemPassive, GlobalSharkAttacks, and PokemonMoveType---
have fewer than 40 distinct normal support points. Public raw data are downloaded from
their official repositories and are not redistributed.

\appendix
\section{Proofs}
\label{app:proofs}

\subsection{Equivalence-class AUROC}
Partition the usual positive--negative pair sum by the equivalence classes of its two
members. There are $p_cn_d$ ordered pairs from classes $(c,d)$ and every such pair has
comparison value $L(s_c,s_d)$, proving Eq.~\eqref{eq:aucdecomp}. When $c=d$, identical
features force identical scores and $L(s_c,s_c)=1/2$, yielding the fixed term.

\subsection{Attained ceiling}
Consider adjacent classes $a,b$ in a proposed score ordering. Ranking $a$ above $b$
contributes $p_an_b$ cross-class wins, whereas the reverse contributes $p_bn_a$. The
first is no smaller iff $p_an_b\geq p_bn_a$, equivalently $q_a\geq q_b$, including
zero-count boundaries. Repeatedly swapping inversions reaches nonincreasing $q_c$)
without reducing the objective. Within-class terms are fixed, so the order is globally
optimal and is attained by the score $q_c$.

\subsection{Canonical replication-invariant weights}
In the separable family, take two positive classes with weights $w^+(r)$ and $w^+(s)$
and one negative class. Choose scores so the first positive wins and the second loses.
After label-wise normalization, the metric contains
$w^+(r)/(w^+(r)+w^+(s))$. Invariance when the first class is replicated from any
$r\geq1$ to any $r'\geq1$, for fixed $s$ with positive weight, forces
$w^+(r)=w^+(r')$. Reversing label roles makes $w^-$ constant on the positive integers.
Counts equal to zero have zero support weight. The normalized metric is therefore
Eq.~\eqref{eq:riauc}; its direct replication invariance follows because the support
indicators do not change.

\subsection{Row-law size bias}
For measurable $A$, the expected number of expanded rows whose support is in $A$ is
proportional to $\int_A\mu(u)dQ(u)$. Dividing by expected total row count
$\int\mu(u)dQ(u)$ gives
$P_{\rm row}(A)=\int_A\mu(u)dQ(u)/\E_Q\mu(U)$ and hence
Eq.~\eqref{eq:sizebias}.

\subsection{Frequency-cause non-identifiability}
For any positive measurable $h(u)$, define $b'(u)=b(u)h(u)$ and
$a'(u)=a(u)/h(u)$. Then $e b'(u)a'(u)=e b(u)a(u)$ for every observed $(u,e)$.
The observable multiplicity mean, and any count law parameterized only through that
product, is unchanged although the causal decomposition differs.

\subsection{Exact replication orthogonality}
Positive replication changes $M_k$ but neither the set nor deterministic order of
distinct representatives. Consequently the class split, support scaler, backbone fit,
calibration scores, and Eq.~\eqref{eq:pvalue} are identical. At scoring time the
backbone evaluates each representative once before mapping the same score to its copies.
Only the explicit count path can change.

\subsection{Factorwise and combined validity}
Conditional on proper training, exchangeability makes the conservative conformal rank
in Eq.~\eqref{eq:pvalue} super-uniform, including ties. Thus each valid factor satisfies
$\Pp(p_j\leq t)\leq t$. The event $p_{\rm comb}\leq\alpha$ implies
$p_S\leq\alpha/2$ or $p_M\leq\alpha/2$. The union bound gives probability at most
$\alpha$ without channel independence. If count exchangeability or exposure
comparability is absent, this argument applies only to $p_S$.

\section{Additional Protocol Notes}

Exact equality is computed after loading the released numeric arrays and canonicalizing
$-0.0$ to $0.0$; NaNs and infinities are audited separately. The audit does not use
approximate hashes. For real-support frequency interventions, dataset-specific random
directions and anomaly support indices are fixed before comparing models. For external
calibration, the held-out normal support is never used for fitting or internal conformal
calibration. Bootstrap resampling always respects the declared unit: dataset for power
and utility, dataset--seed run for external calibration. Pooled Wilson intervals for
individual support tests are produced only as descriptive checks and are not used for
inference.

\bibliographystyle{unsrt}
\footnotesize
\bibliography{references}

\end{document}